\documentclass[conference]{IEEEtran}
\usepackage{pifont}
\usepackage{amsmath}
\usepackage{graphicx}
\usepackage[table]{xcolor}
\definecolor{fucsia}{RGB}{196,0,98}
\usepackage{hhline}

\ifCLASSINFOpdf
\else
\fi
\begin{document}
%
\title{SeLiNet: Sentiment enriched Lightweight Network for Emotion Recognition in Images}

\author{\IEEEauthorblockN{Tuneer Khargonkar\IEEEauthorrefmark{1},
Shwetank Choudhary\IEEEauthorrefmark{2},
Sumit Kumar\IEEEauthorrefmark{3}, 
Barath Raj KR\IEEEauthorrefmark{4}}
\IEEEauthorblockA{ Samsung R\&D Institute, Bangalore, India \\ 
Email: \{\IEEEauthorrefmark{1}t.khargonkar, \IEEEauthorrefmark{2}sj.choudhary, \IEEEauthorrefmark{3}sumit.kr, \IEEEauthorrefmark{4}barathraj.kr\}@samsung.com}
}

\maketitle

\begin{abstract}
In this paper, we propose a sentiment-enriched lightweight network SeLiNet and an end-to-end on-device pipeline for contextual emotion recognition in images. SeLiNet model consists of body feature extractor, image aesthetics feature extractor, and multitask learning-based fusion network which jointly estimates discrete emotion and human sentiments tasks. On the EMOTIC dataset, the proposed approach achieves an Average Precision (AP) score of 27.17 in comparison to the baseline AP score of 27.38 while reducing the model size by $>$85\%. In addition, we report an on-device AP score of 26.42 with reduction in model size  by $>$93\% when compared to the baseline.
\end{abstract}


%
\IEEEpeerreviewmaketitle

\section{Introduction}
\label{sec:introduction}

Understanding the emotional states of the people in the images is an emerging research area in the domain of computer vision. Ability to correctly perceive emotions can help improve human-computer interactions. In the case of smartphones, several use cases can be built such as queries based on-device image search, dynamically uncluttering the notification panel based on user emotions, etc. Further, it has other advanced applications like modeling robot behavior as per the perceived emotion of the user.

Conventionally, researchers have used facial expressions \cite{14, 19} based features to process human emotions. Recently, scientific studies have established that perception of emotions is also influenced by context \cite{3, 4} such as background scene \cite{15, 23}, body posture \cite{17}, image composition \cite{21}, gait analysis \cite{33} etc. Several previous works have achieved better performance by considering these contexts.

Previous studies show that image aesthetics assessment \cite{11} is a crucial cue to understand the emotions evoked by the images. Aesthetics response towards images may depend upon many components such as composition, colorfulness, spatial organization, emphasis, motion, depth, or presence of humans \cite{2, 20}. Traditional works have used low-level handcrafted visual features \cite{1, 10, 31} to understand the aesthetics and related image emotions. Recent works based on deep learning \cite{5, 28, 32} extract mid and global-level features such as composition, semantics, and emphasis \cite{36} to classify image emotions. These works try to understand human emotions evoked by the pictures and are able to achieve improved results by considering the aesthetics properties of the images. Understanding the composition and semantics of the images can help capture the high-level contextual properties like the object’s spatial organization, the relationship between various local level features, etc., and thus can also be beneficial to the task of recognizing the emotional states of people in the images. Image aesthetics assessment has an impact on human sentiment also. It may be either positive, negative, or neutral. For example, images that convey a pleasant mood are generally rated high on the aesthetics scale, and vice-versa. Such images are also known to elicit positive emotions. Taking inspiration from these discussed studies, we explore image aesthetics assessment-based features along with body features to understand the sentiment and emotional states of the people in the images.

Several studies show that privacy is one of the leading concerns among people \cite{22, 24}. With the rise in ownership of smartphones \cite{8}, this concern is particularly high among smartphone users \cite{26, 29}. To this end, we present an end-to-end on-device novel pipeline consisting of the sentiment-enriched lightweight model called SeLiNet for human emotion understanding from the image.
Our main contributions can be summarized as:
\begin{itemize}
\item We propose a sentiment-enriched lightweight model SeLiNet and end-to-end pipeline for on-device emotion recognition in images which achieves comparable average precision(AP) score to the baseline system with a significant reduction in model size (85\% {\textdownarrow }) and inference time(78\% {\textdownarrow }).
\item We demonstrate that the image aesthetics features contribute in improving the overall performance of the task of emotion recognition in images. 
\item We conceptualize the problem as multitask learning-based and make predictions for discrete emotions and related sentiments. We then use sentiment knowledge in the post-processing to enhance emotion predictions and show improved results.
\end{itemize}

\section{Related work}
\label{sec:related_work}

Emotion recognition is a well-studied task in the vision field. Traditional works have used hand-crafted features \cite{25, 37} for the emotion recognition task. Deep learning networks have taken into account facial expressions \cite{14, 19}, gait analysis \cite{33} and body posture \cite{18, 30} etc. based features to predict emotions. \cite{7} proposed facial expressions based on compound emotion detection such as ’happily disgusted’ or ’angrily surprised’ and thus provide deeper insights about expressed emotions. While most of these works have modeled emotion detection tasks as the categorical problem, some have tried to use the valence, arousal, and dominance (VAD) model \cite{27} based on continuous emotional space.

Recently, several works have also demonstrated the importance of contextual cues in interpreting emotional states. \cite{13} presents two-stream encoding networks capturing facial and contextual features, followed by a fusion network to predict context-aware emotion recognition. \cite{12} also proposes similar two-stream architecture in which one stream captures body features and the other stream captures scene context from the image. A fusion network consisting of both body and scene context features is jointly used to predict discrete emotion and VAD scores. They also create and publish the EMOTIC (from EMOTions In Context) Dataset and provide a CNN-based baseline system on the same dataset. \cite{16} proposes context aware multi-modality-based network to predict emotion. They use several context-based modalities such as the face, gait analysis, semantic context, and depth maps to model socio-dynamic interactions among agents. 

In this work, we get inspiration from \cite{12} to design our lightweight network and baseline this work for the comparison of our proposed approach on the EMOTIC dataset. \cite{12} uses resnet50 as a feature extractor for both body and scene context. In contrast, we use the pretrained mobilenetV3\_large model for extracting body features and a lightweight composition-based aesthetics feature extractor (ReLIC \cite{35}) to keep the model size small. This baseline \cite{12} predicts emotion and VAD scores simultaneously. We, instead make sentiment and emotion predictions together and use sentiment predictions in the post-processing module to improve emotion prediction performance. \cite{16} reports an AP score of 35.48 on the EMOTIC dataset. However, they have used several deep neural networks to derive multiple modalities-based contexts, making the overall architecture complex for training and inferencing. Architecture is also computationally intensive, with an overall model size of $>$500MB (includes face detection model size), making it unsuitable for low-resource devices such as smartphones.

We have used the aesthetics feature extractor as one of the branches for our SeLiNet model. Several previous works have shown that there is an explicit connection between image aesthetics and image emotion. \cite{6} have used emotion-assisted image aesthetics identification using multitask learning. They demonstrate that there is a link between image emotion and aesthetics, and that image emotion features can aid in aesthetics assessment tasks and vice versa. \cite{21} use image semantics, image aesthetics, and other visual features to effectively classify the emotion types.

	\begin{figure*}[h ]
		\centering
		\includegraphics[width=\textwidth,height=4cm]{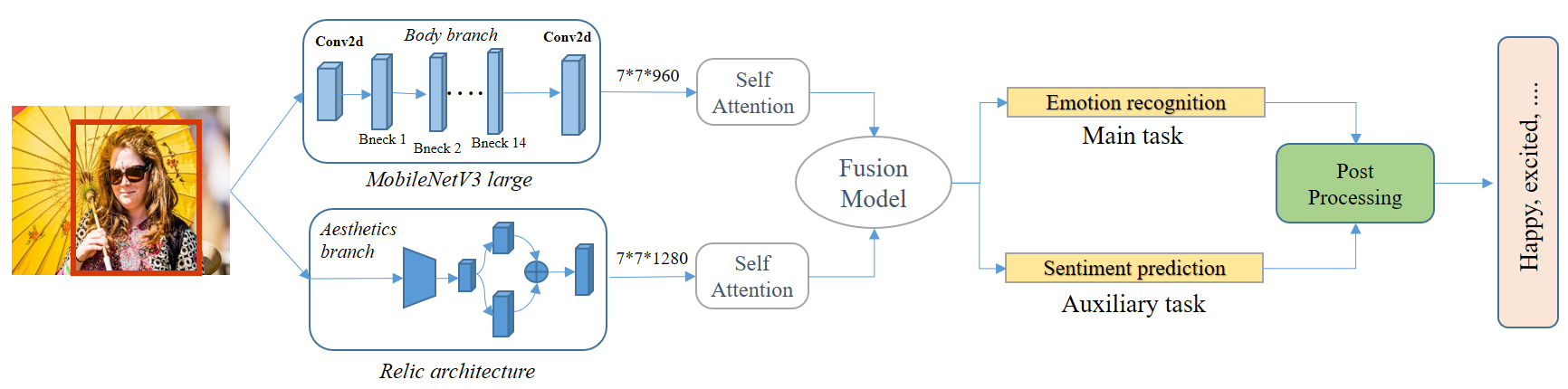}
		\caption{\textbf{Architecture of SeLiNet}}
		\label{fig:model_pipeline}
	\end{figure*}

\section{Dataset}
We train and report our proposed approach performance on EMOTIC \cite{12} dataset. EMOTIC is a benchmark dataset for the context-aware emotion recognition task. It is created by taking images from MSCOCO dataset \cite{38}, Ade20k dataset \cite{39} and images downloaded from the Google search engine, which makes the overall dataset very diverse and increases its complexity. The dataset contains roughly 23,000 images and around 33,000 annotated instances of emotions. The dataset provides bounding box information of the target person in each image and the same has multi-label annotation with 26 possible emotion categories information of the bounding box. Dataset also has annotations for continuous emotion index such as Valance, Arousal, and Dominance (VAD). Emotions are quantified on these three indexes with their scales ranging from 0 to 10. In our work, we have only considered discrete emotions as ground truth and instead added sentiment prediction as an additional task. Since the EMOTIC dataset does not provide sentiment labeling, we use a study by \cite{9} to label the sentiment of each image based on the ground truth emotion. \cite{9} shows each emotion can be categorized into positive, negative, and neutral sentiments. So, we label possible sentiment for each image in the dataset. We use a multi-label strategy to label sentiment because each image can have multiple emotion labels and these labels can fall under more than one sentiment. Table \ref{sentiment_labels} shows a few examples of ground truth emotion labels and corresponding one-hot encoding for the sentiment label. We report all our evaluation results on the test set of the EMOTIC dataset.

\begin{table}[htbp]
\caption{Example of Labeling for Sentiment task }\label{sentiment_labels}
\begin{center}
\begin{tabular}{|c|p{3.5cm}|} 
 \hline
 \cellcolor[HTML]{C0C0C0}{\textbf{Image Emotion Labels}} & \cellcolor[HTML]{C0C0C0}{\textbf{ Sentiment Encoding \newline[Positive,Negative,Neutral] }}  \\ [0.5ex] 
 \hline
 Confidence, Excitement & [1,0,0] \\ 
 \hline
 Peace, Sensitivity & [1,0,1] \\ 
 \hline
 Disapproval, Sadness  & [0,1,0] \\
 \hline
\end{tabular}
\end{center}
\end{table}

\section{The Proposed Method}
This section details the motivation behind the problem and our proposed method.

\subsection{Motivation and Problem Solving}\label{motivation}
Although multi-modality-based information \cite{12, 13, 16} improves the performance of emotion task, the inclusion of these additional information makes the overall model architecture complex \cite{16} and expensive in computation and memory and thus making it unfit for resource constraints systems such as mobile phones. Also, very few works focus on lightweight architectures suitable for the on-device system. Therefore, in this work, we attempted to develop a lightweight model for on-device inferencing. For our method, we derive the idea to employ image aesthetics for the emotion recognition task based on studies discussed in Section \ref{sec:introduction}.

We model the problem in this paper as multitask learning, which predicts both emotions and sentiments. The main idea behind predicting sentiments as an auxiliary task is: 1) To provide an additional loss factor to the emotion task during training in case of incorrect sentiment prediction; 2) To use the sentiment score in post-processing to further enhance the main task (emotion task) performance. It is also possible to infer only the sentiments of an image using the proposed multitask as standalone predictions.

\subsection{The Pipeline}

\subsubsection{Pre-processing}\label{sssec:num1}
The EMOTIC training dataset is highly imbalanced, with certain classes such as engagement, happiness, excitement, etc. occurring more frequently than classes like anger, and aversion. We use standard data augmentation techniques for images such as HorizontalFlip, RandomBrightnessContrast, Posterize, HueSaturationValue, etc. to address the same.

\subsubsection{SeLiNet Model}\label{sssec:num2}
Figure \ref{fig:model_pipeline} shows our SeLiNet architecture and end-to-end pipeline. The proposed SeLiNet model consists of a body module, aesthetics module, and fusion module which are discussed below in detail.

\begin{table}[htbp]
\caption{Performance comparison on the EMOTIC dataset} \label{main_task_comparison}
\begin{center}
\begin{tabular}{|c|c|c|c|} 
 \hline
 \cellcolor[HTML]{C0C0C0}{\textbf{Model}} & \cellcolor[HTML]{C0C0C0}{\textbf{AP Score}} & \cellcolor[HTML]{C0C0C0}{\textbf{Model size}} & \cellcolor[HTML]{C0C0C0}{\textbf{Inference time}} \\ [0.5ex] 
 \hline
 Baseline CNN \cite{12} & 27.38 & $>$190 MB & 16.2 ms \\ 
 \hline
 CAAGRER \cite{34}& 28.42 &  $>$400 MB & NA \\
 \hline
 EmotiCon(GCN) \cite{16}& 32.03 & $>$500 MB & NA \\
 \hline
  EmotiCon(Depth) \cite{16} & 35.48 & $>$500 MB& NA \\
 \hline
 \cellcolor[HTML]{E0FFFF}SeLiNet & \cellcolor[HTML]{E0FFFF}27.17 & \cellcolor[HTML]{E0FFFF}28.03 MB & \cellcolor[HTML]{E0FFFF}3.5 ms \\
 \hline
\end{tabular}
\end{center}
\end{table}


\textbf{Body feature extractor} : This branch focuses on extracting facial and body features from the input image. Extracting these features is important because they provide crucial information about the emotional state of the person in the image. The branch is based on mobilenetV3\_large, trained on the ImageNet dataset with person class. We freeze weights till the second last layer and take its output with a feature map of size (960*7*7) which is then fed to a self-attention network. The attention layer outputs an attentive vector of size 960 which is followed by a dense layer of 512 units whose output is then passed to the fusion model for further processing.

\textbf{Aesthetics feature extractor}: The aesthetics feature extractor uses the pretrained ReLIC architecture \cite{35} as a backbone to extract image aesthetics features. ReLIC architecture \cite{35} is based on several Convolutional Neural Networks and tries to learn both local and global features. Local features are used to understand image composition whereas global features contribute toward the overall image properties such as texture etc. We freeze weights till second last layer and take its output as a feature map of (1280*7*7) and apply self-attention to get attentive feature vectors. The aesthetics branch outputs a 1280-size vector which is followed by a dense layer of 512 units. The output of the dense layer is then fed to the fusion model.

\textbf{Fusion model} : The fusion layer concatenates the output of the body and aesthetics feature extractors to get a 1024-size fused vector. This concatenation layer is then followed by two dense layers of 512 and 256 units. This last 256 dense layer is followed by two task-specific dense layers each of 128 units whose outputs are fed to emotion and sentiment classification layers respectively for the predictions. We perform detailed hyperparameter tuning to choose layers of the fusion model.

\subsubsection{Post-processing}\label{sssec:num3}
We use a boosting algorithm to modify the confidence score of the emotion prediction based on the sentiment output. We consider the top 5 confidence scores by the emotion task $E_i$ as E = {$E_1, E_2, E_3, E_4,E_5$} and predictions by sentiment task $S_j$ as S = {$S_1, S_2, S_3$}. Then the boosting equation is as follows.

\begin{equation}
    E_i = sigmoid(E_i + S_j) ,\; where \; E_i \in S_j
\end{equation}

Our boosting factor $S_j$ provides a relative boost to all emotions in $E_i$ that are predicted correctly in accordance with sentiment output.

\section{Results and Experiments}
This section describes the implementation details, comparison with previous works, and ablation study for our proposed approach.

\begin{table*}[ht]
\caption{Ablation Study} \label{ablation_study}
\begin{center}
\begin{tabular}{|c|c|c|c|c|c|c|}
\hline
 \cellcolor[HTML]{C0C0C0}{\textbf{Body Model}}  & \cellcolor[HTML]{C0C0C0}{\textbf{Aesthetic Model}}  & \cellcolor[HTML]{C0C0C0}{\textbf{ Attention}} & \cellcolor[HTML]{C0C0C0}{\textbf{Sentiment Task}} & \cellcolor[HTML]{C0C0C0}{\textbf{Post Processing}}  & \cellcolor[HTML]{C0C0C0}{\textbf{Model Size}} & \cellcolor[HTML]{C0C0C0}{\textbf{AP Score}}  \\
\hline 
 \ding{51} &  \ding{55} & \ding{55} & \ding{55} & \ding{55} & 12.53 MB & 22.71 \\
 
 \ding{51} &  \ding{51} & \ding{55} & \ding{55} & \ding{55} & 27.23 MB & 25.46  \\
 
\ding{51} &  \ding{51} & \ding{51} & \ding{55} & \ding{55} & 27.91 MB & 26.30 \\

\ding{51} &  \ding{51} & \ding{51} & \ding{51} & \ding{55} & 28.03 MB & 26.81  \\

\cellcolor[HTML]{E0FFFF}\ding{51} &  \cellcolor[HTML]{E0FFFF}\ding{51} & \cellcolor[HTML]{E0FFFF}\ding{51} & \cellcolor[HTML]{E0FFFF}\ding{51} & \cellcolor[HTML]{E0FFFF}\ding{51} & \cellcolor[HTML]{E0FFFF}28.03 MB & \cellcolor[HTML]{E0FFFF}27.17 \\

\hline 
\end{tabular}
\end{center}
\end{table*}

\subsection{Implementation details}
We use the PyTorch framework for experimentation and model development. All training and testing are carried out on an Nvidia GPU GeForce GTX 1080 Ti 11178 MB card. The aesthetics branch takes the complete image of size 224 * 224 * 3 as input. The body branch, on the other hand, requires a 128 * 128 * 3 input image which is a cropped portion of the original image containing the whole body. We set the batch size to 26 and use stochastic gradient descent(SGD) optimizer in the training. The learning rate is initialized to 0.001 with a decay rate of 0.1. The model is trained for 25 epochs and is saved based on the best validation AP score.

\textbf{Loss Function} : Since our problem statement is a multi-class multi-label on the EMOTIC dataset, we experimented with standard binary cross entropy(BCE) loss and L2 loss (suggested by \cite{12}). We observe that L2 loss gives better results than BCE.

\begin{equation}
    loss =  \sum (Y_{(i)actual} - Y_{(i)pred})^2 * W_i
\end{equation}

Where $W_i$ is dynamic weights per batch which are defined as:
    $$
    W_i = 
    \begin{cases}
      \frac{1}{\text{total\; true\; emotions\; present}}, & \text{if emotion is true} \\
      0.0001, & \text{otherwise}
    \end{cases}
    $$

The combined loss of emotion and sentiment task is referred to as the total loss. Based on the experiments, we set $\lambda$ equals to 0.8 which gives better results.
\begin{equation}
    L_{total} = \lambda L_{emotion} + (1- \lambda) L_{sentiment}
\end{equation}

\subsection{Comparison with previous works}
As shown in Table \ref{main_task_comparison}, \cite{12} reports an AP score of 27.38 on the emotion recognition task using a CNN-based baseline system. We try to reproduce their work using the code available on Github \footnote[1]{https://github.com/Tandon-A/emotic}. Using the same configuration discussed in the work, we get an AP score of 25.38 with a model size of nearly 191MB and an inference time of 16.2 ms on GPU. In contrast, using our approach, we achieve an AP score of 27.17 with a model size of only 28.03 MB, a reduction of 85.32\% when compared to the baseline. Our approach is faster by nearly 78\% compared to the baseline. In Table \ref{main_task_comparison}, we also provide a comparison of our proposed model with other works. Although \cite{16} shows better performance, it involves more than three modalities as input to the model making the overall system complex in computation and training and cost-intensive in terms of memory and inference time. In comparison, our work provides for the lightweight model with only two modalities as input and gives comparable performance to the baseline.

For our sentiment sub-task on the EMOTIC dataset, we achieve an AP score of 93.53, 73.82, and 19.13 for Positive, Negative, and Neutral sentiment respectively. Positive and Negative sentiments report better performance compared to Neutral. It is due to the small number of emotions categorized in the neutral sentiment leading to a lower training sample.

\subsection{Ondevice Performance}
 Our end-to-end on-device pipeline is evaluated on a Samsung S21 smartphone (Android SDK 30, 12 GB RAM, 256 GB ROM, Octa-core, Exynos2100 chip). SeLiNet model, quantized using the PyTorch framework, reports an on-device AP score of 26.42 on the same test set of EMOTIC dataset with a model size of 11.34 MB and on-device inference time of 65 ms. Although the AP score drops marginally by nearly 2.76 \% due to quantization, there is a reduction of model size ( of the quantized model) by more than 52\% which is a huge gain. Also, in comparison to the baseline system, \cite{12}, we achieve a comparable AP score by quantized model while reducing the model size by nearly 93\%.

\subsection{Ablation Study}
 Table \ref{ablation_study} describes the ablation study regarding the different configurations of the architecture. The addition of the aesthetics branch results in a nearly 12\% increase in AP score. Our proposed multi-task learning has improved the AP score by at least 3\%. Thus, it demonstrates that all of the components are required to achieve the best results.

\subsection{Error Analysis}
We observe that the performance of our model slightly degrades for the categories such as embarrassment, surprise, yearning, etc. compared to \cite{16}. Possible reasons are 1) The number of original training images is insufficient to learn diverse features. 2) Owing to complex nature of these emotions, additional cues may be required. \cite{16} demonstrates that taking into account multiple modalities leads to better predictions.

The difficult nature of the emotion recognition task and the requirement of multiple contexts are supported by the study conducted by \cite{16} where if only the facial context is considered out of three discussed contexts, then the AP score drops from 35.48 to 24.06. The baseline by \cite{12} has used Resnet50 for both body and scene context feature extraction. But, when we replace Resnet50 with ResNet18, the AP score falls to 17.23, indicating that shallower models are insufficient for the task. Considering the above details, SeLiNet performs fairly well despite being lightweight.

\subsection{Conclusion}
We present a lightweight model SeLiNet and an end-to-end pipeline to predict the emotional states of people in images for on-device inferencing. Our proposed approach achieves an AP score of 27.17, which is comparable to the baseline system with a 85\% smaller memory footprint and much faster inference time. We also show that aesthetics assessment of the images can be helpful information to understand image emotion. Using multitasking learning, we further improve our model results. In future work, we will like to capture additional contextual information such as object detection, deeper semantic analysis, and its impact on image emotion recognition tasks.

\end{document}